\documentclass{article}
\usepackage{spconf,amsmath,epsfig,amssymb,multirow}
\usepackage{enumitem}
\let\OLDthebibliography\thebibliography
\renewcommand\thebibliography[1]{
  \OLDthebibliography{#1}
  \setlength{\parskip}{0pt}
  \setlength{\itemsep}{0pt plus 0.3ex}
}

\begin{document}\sloppy

% Example definitions.
% --------------------
\def\x{{\mathbf x}}
\def\L{{\cal L}}

% Title.
% ------
\title{SELF-SUPERVISED VIDEO REPRESENTATION LEARNING WITH MOTION-CONTRASTIVE PERCEPTION}
%
% Single address.
% ---------------
% \name{Anonymous ICME submission}

\name{Jinyu Liu\textsuperscript{1}, Ying Cheng\textsuperscript{2}, Yuejie Zhang\textsuperscript{1}, Rui-Wei Zhao\textsuperscript{2}, Rui Feng\textsuperscript{1, 2, *}\thanks{*Corresponding author.}}
\address{\textsuperscript{1}School of Computer Science, Shanghai Collaborative Innovation Center of Intelligent Visual Computing,\\ Fudan University, China
\\\textsuperscript{2}Academy for Engineering and Technology, Fudan University, China
\\\{jinyuliu20, chengy18, yjzhang, rwzhao, fengrui\}@fudan.edu.cn}

%Address and e-mail should NOT be added in the submission paper. They should be present only in the camera ready paper. 
\maketitle

\begin{abstract}
Visual-only self-supervised learning has achieved significant improvement in video representation learning. Existing related methods encourage models to learn video representations by utilizing contrastive learning or designing specific pretext tasks. However, some models are likely to focus on the background, which is unimportant for learning video representations. To alleviate this problem, we propose a new view called \emph{\textbf{long-range residual frame}} to obtain more motion-specific information. Based on this, we propose the \textbf{M}otion-\textbf{C}ontrastive \textbf{P}erception \textbf{Net}work (\textbf{MCPNet}), which consists of two branches, namely, \textbf{M}otion \textbf{I}nformation \textbf{P}erception (\textbf{MIP}) and \textbf{C}ontrastive \textbf{I}nstance \textbf{P}erception (\textbf{CIP}), to learn generic video representations by focusing on the changing areas in videos. Specifically, the MIP branch aims to learn fine-grained motion features, and the CIP branch performs contrastive learning to learn overall semantics information for each instance. Experiments on two benchmark datasets UCF-101 and HMDB-51 show that our method outperforms current state-of-the-art visual-only self-supervised approaches.

\end{abstract}

\section{Introduction}
\label{sec:intro}
Human brains with a strong understanding ability can process visual information as seen by eyes and quickly infer its meaning. There are a lot of video understanding tasks, such as video segmentation and video event localization. The most significant difference between video and image understanding is that videos contain complex spatial-temporal contents, but pictures only have spatial information. Hence, how to learn effective video representation is essential yet challenging. However, large-scale datasets require laborious and expensive annotation such as Kinectis-400~\cite{kinetics}, etc. Due to large-scale unlabelled data on the Internet, video self-supervised representation learning methods have attracted great attention. 

\begin{figure}[t]
    \centering
    \includegraphics[width=0.4\textwidth]{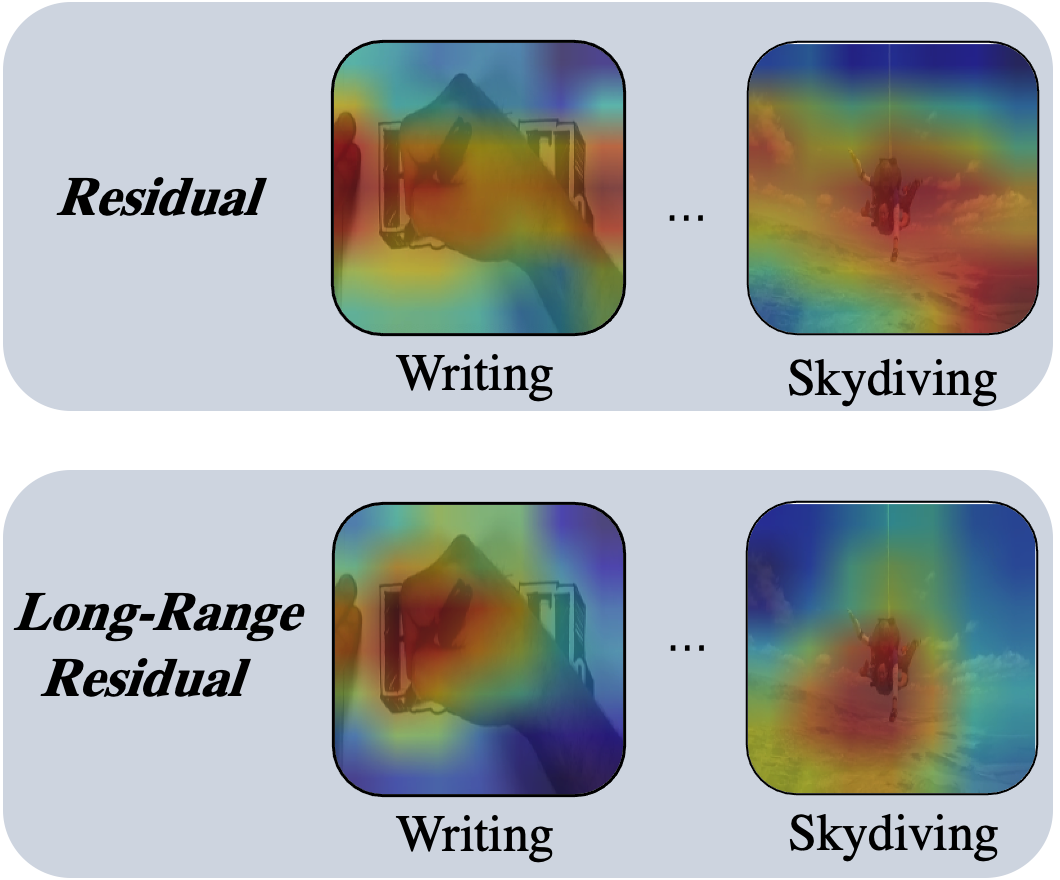}
    \caption{\textbf{Comparison between models trained with residual frames and with long-range residual frames on Kinetics-100~\cite{rspnet}.} The above activation maps are produced by the last convolutional layer of the S3D-G backbone. By utilizing long-range residual frames as one of the inputs in our proposed method, the learned representations can capture motion areas more accurately.}
    \label{figure1}
\end{figure}
Self-supervised video representation learning methods aim to learn helpful information through pretext tasks that leverage supervision in the data itself and significantly reduce the cost of collecting manual labels. In recent years, some efforts have been made in self-supervised video representation learning. The pretext tasks in these works can be divided into three categories: 1) spatial-focused learning, such as geometry guided learning \cite{geometry}; 2) temporal-focused learning, such as frame sequencing \cite{shuffle, sequences}, clip orders prediction \cite{cliporder} and playback rate perception \cite{prp, pace, speednet, rspnet}; 3) spatial-temporal learning, such as space-time cubic puzzles \cite{space-time-cubic} and multi-task \cite{multitask}. However, a problem with most of these pretext tasks is that the information contained in the RGB view is redundant. Some stationary and semantically irrelevant objects in the background are likely to interfere with the model making judgments. Some researchers \cite{removingbackground, context-motion-decoupling} point out that poor video comprehension is primarily the result of background cheating. Therefore, some latest methods utilize extra views such as optical flow \cite{coclr}, residual frame \cite{iic}, etc., to encourage models to focus on the changing areas rather than the irrelevant part of the background, which has achieved convincing improvement. However, the calculation of optical flow is generally expensive, and optical flow is sensitive to light changes. Thus residual frame that requires cheap computation is a more reasonable view, and we propose \emph{\textbf{long-range residual frame}}, as shown in Fig.~\ref{figure1}, to obtain more motion-specific information. 

In this paper, we propose a novel \textbf{M}otion-\textbf{C}ontrastive \textbf{P}erception \textbf{Net}work (\textbf{MCPNet}), which consists of two branches, namely, \textbf{M}otion \textbf{I}nformation \textbf{P}erception (\textbf{MIP}) and \textbf{C}ontrastive \textbf{I}nstance \textbf{P}erception (\textbf{CIP}). For the MIP branch, which aims to learn fine-grained motion features, we sample an RGB clip and a long-range residual clip from the same video, requiring the model to distinguish whether the playback speeds of these two clips are the same. For the CIP branch, which is designed to learn overall semantics information for each instance, we encourage the representations of an RGB clip and a long-range residual clip sampled from the same video with different playback speeds to be close enough in feature space. Two branches are trained jointly during pre-training. Experimental results on two datasets show that the learned features perform well on two downstream tasks, i.e., action recognition and video retrieval.

To summarize, the contributions of this paper are as follows:

\begin{itemize}[leftmargin=*]
\item We propose a simple-yet-effective view called long-range residual frame for self-supervised video representation learning, which contains more motion-specific information. 
\item We propose a novel Motion-Contrastive Perception Network (MCPNet), consisting of a MIP branch and a CIP branch, encouraging the model to focus more on the moving objects and less on the static and irrelevant objects in the background.
\item Experiments show that our model can achieve state-of-the-art results for two downstream tasks of action recognition and video retrieval on both two benchmark datasets UCF-101 and HMDB-51.
\end{itemize}

\section{Related Work}
\noindent\textbf{Contrastive Learning}. Contrastive learning has achieved a lot of success in self-supervised learning \cite{pace, iic, coclr, rspnet, ascnet}, which does not pay too much attention to pixel details but can focus on abstract semantic information. Chen et al.~\cite{simple-framework} proposed a simple framework without requiring specialized architectures or memory bank, which enables the contrastive tasks to learn useful representations. Tian et al.~\cite{cmc} presented a contrastive multi-view coding approach for video representation learning, which used different views of input videos to maximize the instance-level distinction. He et al.~\cite{momentum} proposed MoCo, which could build a large and consistent dictionary with a queue that could enqueue and dequeue learned embeddings and a moving-averaged encoder that maintained consistency. Their work facilitates contrastive unsupervised learning.

\noindent\textbf{Self-Supervised Visual Representation Learning}. Early visual self-supervised representation learning methods mainly focused on images, which designed pretext tasks such as colorization \cite{colorization}, jigsaw puzzles \cite{jigsaw-puzzles}, image rotations \cite{rotation}, relative position \cite{contextprediction}, etc. Later some self-supervised approaches for videos emerged, such as space-time cubic puzzles \cite{space-time-cubic}, frame order prediction \cite{shuffle, sequences}, clip order prediction \cite{cliporder}, multi-task~\cite{multitask}, etc. Recently, playback speed perception \cite{prp, pace, speednet, rspnet} has attracted a lot of attention, which requires the model have a deep understanding of video contents to accomplish tasks. Some latest works used additional views such as optical flow~\cite{coclr}, residual frame~\cite{iic}, etc. However, many pretext tasks ignore the background cheating problem in RGB views. Some approaches use additional views to improve the performance of models but do not have a carefully designed pretext task. Our method makes full use of long-range residual frames, relative speed perception, and contrastive learning to learn generic video representations.

\section{Method}

\begin{figure*}[t]
\centering
\includegraphics[width=0.95\textwidth]{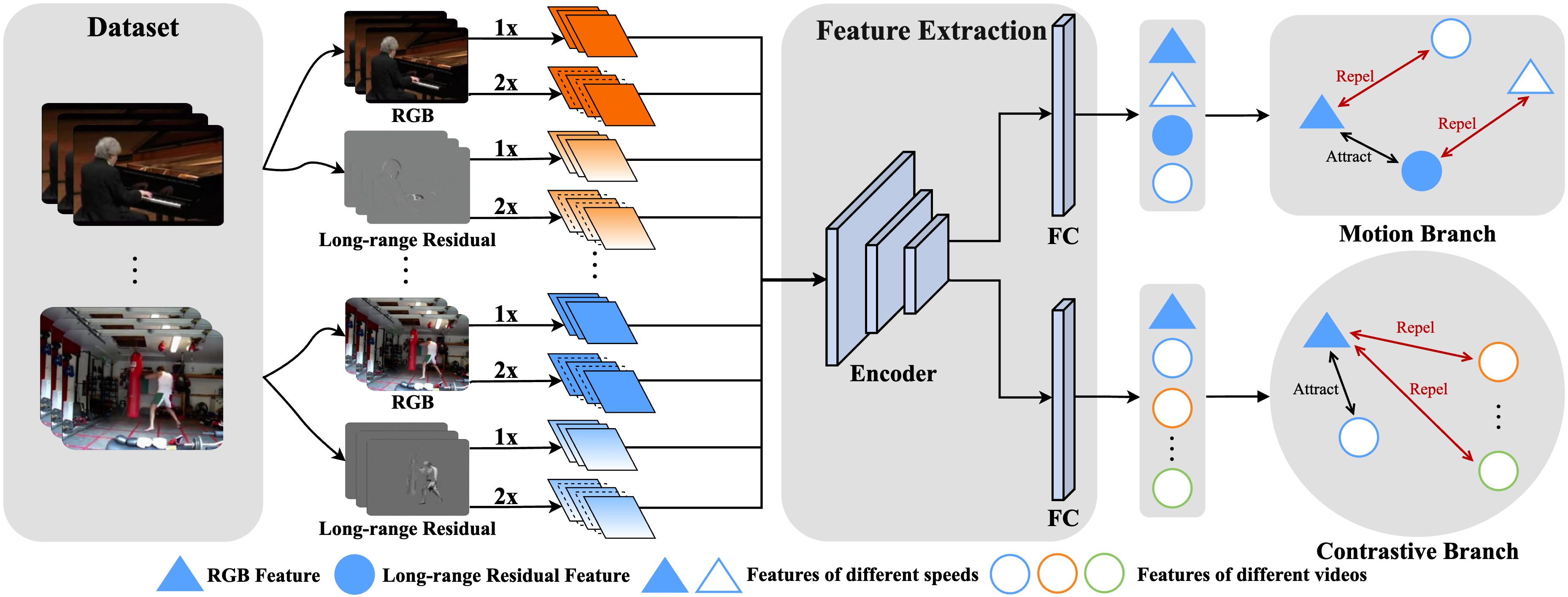}
\vspace{-3mm}
\caption{An overview of our Motion-Contrastive Perception Network. Our method consists of two branches: Motion Information Perception (MIP) and Contrastive Instance Perception (CIP). MIP branch learns fine-grained motion features by distinguishing the speed difference between two different view clips. CIP branch performs contrastive learning by predicting whether two different view clips come from the same video to learn general semantics information for each instance.}
\label{figure2}
\vspace{-4mm}
\end{figure*}

\subsection{Long-Range Residual Frames}
Multi-view inputs have been proved to be efficient for instance-based video contrastive learning. Optical flow~\cite{iic, coclr} contains rich motion information but is computationally intensive. The residual frame was first employed by Tao et al.~\cite{residual} in video representation learning, which could save frame difference with much lower computational volume compared to optical flow. Tao et al.~\cite{iic} demonstrated that stacked residual frames were also very effective in self-supervised video representation learning. The calculation of the residual frame is shown as below:

\vspace{-15pt}
\label{res}
\begin{eqnarray}
   Res_{i\sim j} = | Frame_{i\sim j} - Frame_{i+1\sim j+1} |,
  \vspace{-5pt}
\end{eqnarray}
where $Res$ represents residual frame, $Frame$ represents RGB frame, $i\sim j$ is the index interval of the sampled frames.

The common aim of the residual frame and optical flow is to encourage the model to focus on the changing areas of videos rather than the stationary parts. We observe that the variation between adjacent frames is relatively small, while the variation between two frames farther away in the time dimension is huge. Based on this observation, we propose the \emph{\textbf{long-range residual frame}}, which is generated by selecting two frames 
with a suitable interval in the time dimension to make a difference. Compared to the residual frame, the long-range residual frame contains more frame difference information with the same amount of computation but does not add too much interference information. The process to get long-range residual frames can be formulated as:

\vspace{-15pt}
\begin{eqnarray}
\label{longres}
   LongRes_{i\sim j} = | Frame_{i\sim j} - Frame_{i+t\sim j+t} |,
  \vspace{-5pt}
\end{eqnarray}
% where $LongRes$ represents long-range residual frame.

With stacked long-range residual frames, the movement preserved covers greater information both in spatial and temporal dimensions compared to stacked residual frames. Therefore, models can extract more specific motion features by focusing on the movements in videos.

\subsection{Motion-Contrastive Perception Network}
We propose the Motion-Contrastive Perception Network, which consists of two branches: motion information perception and contrastive instance perception, as shown in Fig.~\ref{figure2}.

\noindent\textbf{Motion Information Perception.}
MIP branch aims to capture fine-grained motion features by distinguishing the speed difference between two different view clips. Let $V = \{ v_i \}_{i=1}^{N}$ be a video set containing $N$ videos. Given a video $v_i$, we sample an RGB clip $r_i$ and two long-range residual clips $l_i^1$, $l_i^2$ with playback speeds $s_r$, $s_{l1}$ and $s_{l2}$,  respectively, where $ s_r = s_{l1} \neq s_{l2}$. It needs to be mentioned that an accelerated RGB clip is obtained by taking frames at intervals (e.g., for a 2x playback speed RGB clip, sampling interval is set as 2 frames), and the same playback speed long-range residual clip can be generated by inputting the accelerated RGB clip into Eq. (\ref{longres}) for calculation. We feed the clips into the video encoder $e(\cdot;\theta)$ followed by a projection head $g_m(\cdot;\theta_m)$ to obtain the features $f_i^r$, $f_i^{l1}$ and $f_i^{l2}$. We use triplet loss~\cite{triplet-loss} as the loss function of MIP, which can be formulated as:

\vspace{-15pt}
\begin{equation}
\label{loss_m}
    {{{\cal L}}_{mip}} = { {\rm max}(\gamma - (sim(f_i^r, f_i^{l1}) - sim(f_i^r, f_i^{l2})), 0) },
    \vspace{-5pt}
\end{equation}
where $\gamma > 0$ is a certain margin and set to 2.0, $sim(, )$ is a dot product function to measure the similarity between two features. The similarity of a positive pair $\{f_i^r, f_i^{l1}\}$ should be larger than a negative pair $\{f_i^r, f_i^{l2}\}$ by a margin $\gamma$. 

\noindent\textbf{Contrastive Instance Perception.}
Contrastive learning aims to distinguish different instances from feature space to gain abstract semantics information. In the CIP branch, we ensure that the speeds of RGB view and long-range residual view are always different, thus encouraging the model to pay attention to the overall semantic information for each instance. Two different views of the same video $v_i$, \textit{e.g.}, $\{$$r_i, l_i$$\}$, are treated as positive, while the views from different videos, \textit{e.g.}, $\{$$r_i, l_j$$\}$ $(i$$\not=$$j)$, are regarded as negative. Specifically, we sample an RGB clip $r_i$ from $v_i$ and $N$ long-range residual clips $\{l_n\}_{n=1}^{N}$ from $V$. Then we feed each clip into the encoder $e(\cdot;\theta)$ followed by a projection head $g_c(\cdot;\theta_c)$ to obtain the corresponding features $f_i^r$ and $F^l$ $=$ $\{$$f_1^l$, ... , $f_i^l$, ..., $f_{n}^l$$\}$. The feature set $F^l$ consists of one positive sample $f_i^l$ and $n-1$ negative samples. We use InfoNCE loss~\cite{info-nce} as our CIP loss, which can be formulated as:

\vspace{-15pt}
\begin{align}
\label{loss_c}
    {{{\cal L}}_{cip}} = {-log{\frac{exp(sim({f_{i}^{r}, f_{i}^{l}})/{\tau})}
    {\sum_{j=1}^{n}exp(sim({f_{i}^{r}, f_{j}^{l}})/{\tau})}}},
    \vspace{-5pt}
\end{align}
where $n$ is the number of negative samples, $f_{i}^{r}$ and $f_{i}^{l}$ are extracted features of two different views from $ith$ video. ${\tau}$ is a temperature hyper-parameter.
% which affects the concentration level of distribution.

\noindent\textbf{Optimization.} MIP loss and CIP loss from the two branches are combined to get the final loss, which is defined as:

\vspace{-15pt}
\begin{align}
    {{{\cal L}_{final}}} = \alpha \cdot {{\cal L}_{mip}} + (1-\alpha) \cdot {{\cal L}_{cip}},
    \vspace{-5pt}
\end{align}
where $\alpha$ = 0.5 is a fixed hyper-parameter to control the importance of each term. MIP branch and CIP branch are pre-trained jointly, and losses ${{\cal L}}$, ${{\cal L}_{mip}}$ and ${{\cal L}_{cip}}$ are used to optimize model parameters $\theta$, $\theta_m$ and $\theta_c$ with gradient descent.
	
\section{Experiments}
\subsection{Experimental Setup}
\noindent\textbf{Datasets.}
In our experiments, we consider four video datasets, namely, Kinetics-400~\cite{kinetics}, Kinetics-100~\cite{rspnet}, UCF-101~\cite{ucf101}, and HMDB-51~\cite{hmdb51}. The Kinetics-400~ dataset contains 400 human action categories and provides approximately 240k training video clips and 20k validation video clips. For the self-supervised pre-training, we use the training split of the Kinetics-400 by discarding all of the labels. Each sample is temporally trimmed to be approximately 10 seconds. In ablation studies, to reduce training costs, we use the Kinetics-100 dataset, which consists of 100 classes with the least disk size of videos from Kinetics-400. 

For the downstream tasks, UCF-101 and HMDB-51 are used to evaluate the effectiveness of our method. UCF-101 consists of 13,320 videos from 101 human action classes, and HMDB-51 contains about 7K video clips of 51 different human motion classes.

\noindent\textbf{Backbones.} We explore three different backbone networks as the video encoder in ablation studies, \textit{i.e.}, R3D-18~\cite{resnet}, R(2+1)D~\cite{r(2+1)d}, and S3D-G~\cite{s3dg}. For action recognition task, the results of R3D-18 and S3D-G are reported. For video retrieval task, the result of R3D-18 is reported.

\noindent\textbf{Self-supervised Pre-training.} We sample 16 frames with 112×112 spatial size for each clip unless specified otherwise. The possible playback speed $s$ for clips is set to 1x and 2x. We also use random cropping with resizing, horizontal flips, and color jittering for augmentation. The SGD algorithm is used to optimize our model. We set the initial learning rate to 0.1, scaled linearly with the batch size $b$, i.e., the learning rate is set to 0.1×$b$/32. The batch sizes of S3D-G and R3D-18 are 28 and 256, respectively. We train our models using 4 NVIDIA Quadro RTX 6000 for 200 epochs.

\noindent\textbf{Fine-tuning.} We initialize the models with the weights from the pre-trained MCPNet, and added two fully-connected layers with randomly initialized weights for classification. We fine-tune the entire model on UCF-101 and HMDB-51 with a learning rate of 0.005 for the action recognition task.

\noindent\textbf{Evaluations.} To evaluate the generalizability and transferability of our proposed method, we apply the pre-trained model to action recognition and video retrieval tasks. For action recognition, the top-1 accuracies on UCF-101 and HMDB-51 are reported. For video retrieval, top-1, top-5, top-10, top-20, and top-50 accuracies are compared with existing approaches.

\subsection{Ablation Studies}
\label{ablation}
We conduct ablation experiments on the influence of $t$, the effectiveness of individual branch, and the effectiveness of long-range residual frames, respectively. All models are pre-trained with 200 epochs on the Kinetics-100 dataset, except for the w/o pre-training setting.

\noindent\textbf{Influence of $t$.} As depicted in Eq. (\ref{longres}), $t$ is the suitable interval between two frames for long-range residual frame generation. We conduct experiments to explore the influence of this hyper-parameter. Table~\ref{table1} shows the results of 5 settings of $t$ with S3D-G backbone. The setting of $t = 1$ means that the residual frames are used in our method. It can be observed that the best result is obtained when $t$ = 4, so we will set $t$ = 4 for all subsequent experiments.
\begin{table}[h]
    \centering
    \caption{Results of different $t$ settings on action recognition.}
	\begin{tabular}{|c|c|c|}
	\hline
	    \textbf{Settings}  & \textbf{UCF-101(\%)} & \textbf{HMDB-51(\%)}  \\ \hline
		$t$ = 1  &  70.2 & 40.3\\ 
		$t$ = 2  &  70.9 & 40.8\\
		$t$ = 3  &  71.6 & 41.2\\ 
		$t$ = 4  & \textbf{72.3} & \textbf{41.7}\\
		$t$ = 5  & 71.1 & 41.4\\ \hline
	\end{tabular}   
	\label{table1}
\vspace{-3mm}
\end{table}	

\noindent\textbf{Effectiveness of Individual Branch.} To figure out the contributions of each branch to the final performance, we conduct ablation studies on six ablated models with the S3D-G backbone built upon our base model. The results of action recognition on UCF-101 are shown in Table~\ref{table2}. Compared to training from scratch, pre-training with only the MIP branch can significantly improve the performance from 45.30\% to 68.44\% on the UCF-101 dataset. Meanwhile, when combing CIP and MIP, we also investigate the speed sampling configuration for the CIP branch. The RGB clips and the long-range residual clips used in the CIP branch can be sampled with the same speeds, random speeds, or different speeds. When the speeds of RGB clips and long-range residual clips are always different, combining CIP and MIP further improves the performance from 68.44\% to 72.35\%, indicating the effectiveness of cooperative work of the proposed two branches.

\begin{table}[h]
    \centering
    \caption{Ablation study for different components of MCPNet on action recognition.}
    	\begin{tabular}{|c|c|c|}
        	\hline
        	\textbf{Method} & \textbf{Configuration} & \textbf{UCF-101(\%)} \\ 
            \hline
            w/o pre-training & - &45.30  \\
            w/ CIP only & different speed &64.87 \\ 
            w/ MIP only & - &68.44 \\
            MIP + CIP & random speed &70.46\\
            MIP + CIP & same speed &70.74\\
            \hline
            MIP + CIP (Ours) & different speed &\textbf{72.35} \\ 
        	\hline
    	\end{tabular}  
	\label{table2}
\end{table}	

\begin{table}[ht]
    \centering
    \caption{Comparisons between residual view and long-range residual view of different backbones for action recognition accuracy (\%) on the UCF-101 dataset.}
    	\begin{tabular}{|c|c|c|c|}
        	\hline
        	\textbf{Backbone}  & \textbf{Random} & \textbf{Residual} & \textbf{LongRes} \\
            \hline
            R3D-18 & 42.4 & 67.8 & \textbf{69.1}\\
            R(2+1)D & 56.0 & 78.2 & \textbf{79.0}\\
            S3D-G & 45.3 & 70.2 & \textbf{72.3}\\ 
        	\hline
    	\end{tabular}  
	\label{table3}
\end{table}

\begin{table*}[t]
% \vspace{-5mm}
    \centering
    \caption{Comparisons with state-of-the-art methods for action recognition accuracy (\%) on UCF-101 and HMDB-51 datasets.}
    	\begin{tabular}{|l|ccc|cc|}
    	    \hline
    	    \textbf{Methods}  & \textbf{Pre-train Dataset} & \textbf{Backbone}& \textbf{Resolution} & \textbf{UCF-101} & \textbf{HMDB-51}  \\ \hline
    		Shuffle\&Learn~\cite{shuffle}&UCF-101 & CaffeNet & 224 & 50.2 & 18.1   \\ 
    		CMC~\cite{cmc} & UCF-101 & CaffeNet & 224 & 59.1 & 26.7 \\
    		VCP~\cite{vcp} & UCF-101 & R(2+1)D & 112 & 66.3 & 32.2 \\
    		ClipOrder~\cite{cliporder}&UCF-101& R(2+1)D & 112 & 72.4 & 30.9  \\ 
    		PRP~\cite{prp}&UCF-101& R(2+1)D & 112 & 72.1 & 35.0  \\ 
    % 		DPC~\cite{dpc}& UCF-101     & R3D-18  & 128 & 68.2   & 34.5   \\
		    IIC~\cite{iic}& UCF-101 & R3D-18 & 112 & 74.4& 38.3\\
            3D-RotNet~\cite{3DRotNet} &Kinetics-400 & R3D-18    & 112 & 62.9 & 33.7   \\
    		ST-Puzzle~\cite{space-time-cubic} &Kinetics-400& R3D-18  & 224 & 63.9 & 33.7   \\
		    DPC~\cite{dpc} &Kinetics-400& R3D-18 & 128 & 68.2 & 34.5 \\
		  %  MCN~\cite{mcn} &Kinetics-400& R3D-18 & 128 & 89.7 & 59.3 \\
    		SpeedNet~\cite{speednet}&Kinetics-400 & S3D-G & 224 & 81.1 & 48.8  \\	
    		Pace~\cite{pace} &Kinetics-400 & S3D-G & 224 & 87.1 & 52.6  \\
    		CoCLR~\cite{coclr} &Kinetics-400& S3D-G & 128 & 87.9 & 54.6   \\	
    		RSPNet~\cite{rspnet} &Kinetics-400 & S3D-G & 224 & 89.9 & 59.6   \\	
    		ASCNet~\cite{ascnet} &Kinetics-400 & S3D-G & 224 & 90.8 & 60.5   \\	
    		\hline
            MCPNet (Ours)&Kinetics-400 & R3D-18 & 112 & 82.2 & 52.5\\
            MCPNet (Ours)&Kinetics-400 & S3D-G & 224 & \textbf{91.5} & \textbf{62.6}\\
    		\hline
    	\end{tabular}
    % }
    \label{table4}
\end{table*}

\noindent\textbf{Effectiveness of Long-Range Residual Frames.} We conduct ablation studies on both residual view and long-range residual view. The results of the action recognition task with different video encoders on UCF-101 are illustrated in Table~\ref{table3}. Compared to the residual frames, the model with long-range residual frames consistently improve 1.3\%, 0.8\%, and 2.1\% on R3D-18, R(2+1)D, and S3D-G, respectively. The comparison results demonstrate the effectiveness of the long-range residual frames.

\subsection{Evaluation on Action Recognition Task}
We compare the results of fine-tuning all parameters with other state-of-the-art methods. Specifically, we pre-train our models on Kinetics-400 and then fine-tune the pre-trained models on UCF-101 and HMDB-51. For S3D-G, we use video frames with 224 × 224 spatial size as input for pre-training and fine-tuning to exploit the proposed approach's potential further. Considering that long-range residual frames and RGB frames are both RGB information in our experiments, we only include the RGB-only results of CoCLR for a fair comparison. From Table~\ref{table4}, we can observe that our models with both S3D-G and R3D-18 backbones outperform other state-of-the-art self-supervised approaches. 

\subsection{Evaluation on Video Retrieval Task}

To further verify the effectiveness of our MCPNet, we also evaluate it on video retrieval task, which can better reflect the semantic-level learning capability. RGB views of original video clips are considered for video retrieval. We directly extract features from the pre-trained model with R3D-18 backbone for video retrieval without fine-tuning. Given the visual feature of a video from the test set as query, video retrieval task aims to return k nearest videos from the training set. When the class of retrieval video is the same as that of the query video, this retrieval result is considered correct. The top-1, top-5, top-10, top-20, and top-50 retrieval accuracies have been shown in Table~\ref{table5}. Compared to other state-of-art methods, our method achieves competitive performance on the UCF-101 dataset. We observe that the top-1 accuracy of CoCLR is better than ours. However, CoCLR uses the optical flow that is computationally complex with extremely accurate motion information as input and adopts dual models. At the same time, we only adopt one single model and use the long-range residual frame with negligible computational cost.

% but MCPNet can take the model trained by CoCLR as a baseline to further pretrain and take additional improvement.
\begin{table}[h]
	\centering
	\caption{Comparisons with previous methods for video retrieval task on the UCF-101 dataset.}
	\resizebox{\columnwidth}{!}{
    	\begin{tabular}{|l|c|c|c|c|c|}
    		\hline
    		\textbf{Method} & \textbf{Top-1} & \textbf{Top-5} & \textbf{Top-10} & \textbf{Top-20} & \textbf{Top-50}    \\ 
    		\hline
    		SpeedNet~\cite{speednet}  & 13.0 & 28.1 & 37.5 & 49.5 & 65.0   \\ 
    		ClipOrder~\cite{cliporder}  & 14.1 & 30.3 & 40.0 & 51.1 & 66.5   \\
    		Jigsaw~\cite{jigsaw-puzzles} & 19.7 & 28.5 & 33.5 & 40.0 & 40.9\\
    		OPN~\cite{sequences} & 19.9 & 28.7 & 34.0 & 40.6 & 51.6   \\
    		Buchler~\cite{buchler2018improving} & 25.7& 36.2& 42.2& 49.2& 59.5 \\
    		VCP~\cite{vcp}  & 18.6   & 33.6     & 42.5   & 53.5     & 68.1 \\
    		CMC~\cite{cmc} & 26.4 & 37.7 & 45.1 & 53.2 & 66.3 \\
    		Pace~\cite{pace} & 31.9 & 49.7 & 59.2 & 68.9 & 80.2   \\ 
    		IIC~\cite{iic} & 42.4 & 60.9  & 69.2 & 77.1 & 86.5 \\
    	    RSPNet~\cite{rspnet}   & 41.1  & 59.4 & 68.4 & 77.8  & 88.7 \\ 
    	    CoCLR~\cite{coclr} & \textbf{53.3} & 69.4 & 76.6 & 82.0 & -\\
    	   % MCN~\cite{mcn} &53.8 & 70.2 &  78.3 &  83.4 & 89.7  \\
    		\hline
    		MCPNet (Ours)  & 48.5 & \textbf{71.6}  & \textbf{78.8} & \textbf{86.0}  & \textbf{92.8}  \\ 
    		\hline
    	\end{tabular}
	}
	\label{table5}
\end{table}

\begin{figure}[h]
    \centering
    \includegraphics[width=0.47\textwidth]{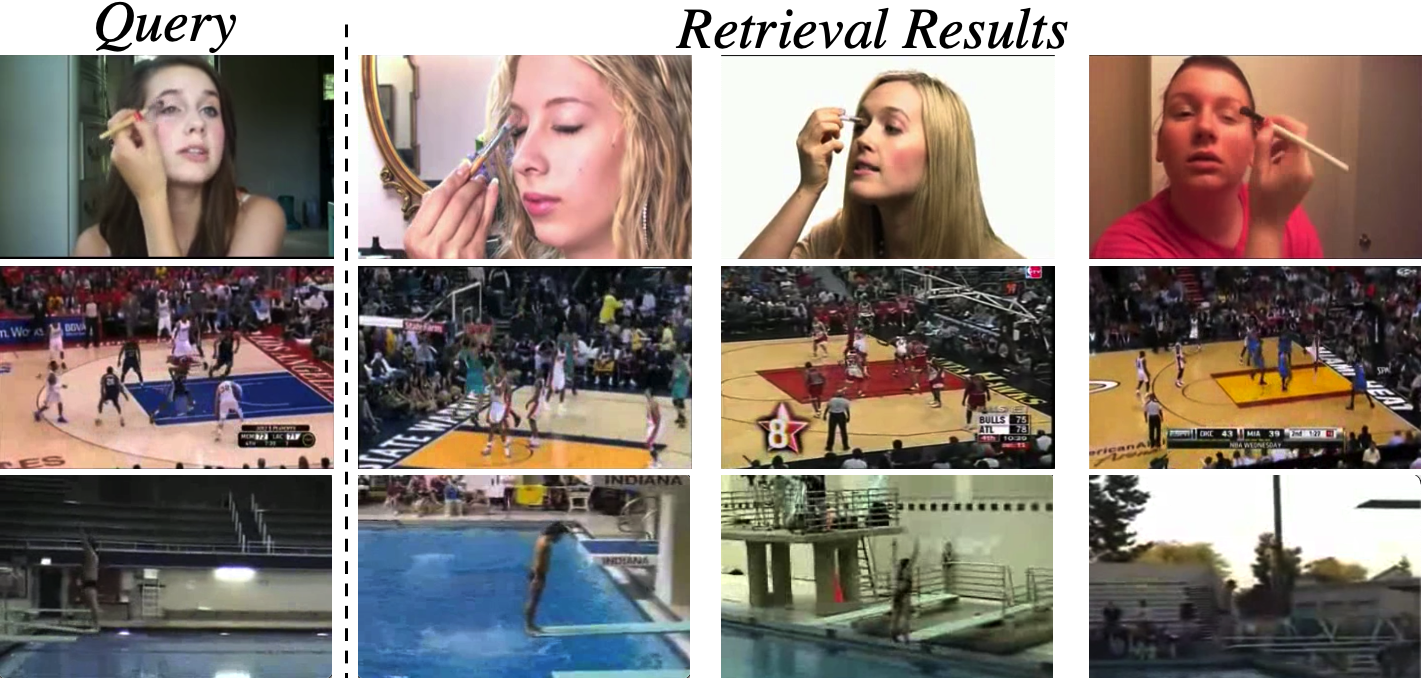}
    \vspace{-3mm}
    \caption{Qualitative examples of the video retrieval task.}
    \label{figure3}
\end{figure}

To qualitatively assess the capabilities of our model, we have given some examples in Fig.~\ref{figure3}. The left is the query video from the UCF-101 testing set, and the right shows the top-3 nearest neighbors from the UCF-101 training set. It can be seen that our method can accurately retrieve videos of the same category.

\section{Conclusion}
\vspace{-1mm}
In this paper, we propose a novel \textbf{M}otion-\textbf{C}ontrastive \textbf{P}erception \textbf{Net}work (\textbf{MCPNet}), consisting of two branches. One is the Motion Information Perception (MIP), which learns fine-grained motion features by distinguishing the speed difference between two different view clips, and the other is the Contrastive Instance Perception (CIP) to learn overall semantics information for each instance by distinguishing whether two different view clips come from the same video. By utilizing the proposed stacked long-range residual frames produced by original RGB frames, that is only using information from the RGB frames, our method outperforms state-of-the-art visual-only self-supervised methods on action recognition and video retrieval tasks. 

\section{Acknowledgments}
\vspace{-3mm}
This work was supported by the National Natural Science Foundation of China (No. 62172101, No. 61976057), the Science and Technology Commission of Shanghai Municipality (No. 20511100800, No. 21511100500), and the Science and Technology Major Project of Commission of Science and Technology of Shanghai (No. 2021SHZDZX0103).

% -------------------------------------------------------------------------

\begin{small}
\bibliographystyle{IEEEbib}
\bibliography{reference}
\end{small}
\end{document}